\documentclass{article}

\usepackage{arxiv}

\usepackage[utf8]{inputenc} 
\usepackage[T1]{fontenc}    
\usepackage{hyperref}       
\usepackage{url}            
\usepackage{booktabs}       
\usepackage{amsfonts}       
\usepackage{nicefrac}       
\usepackage{microtype}      
\usepackage{lipsum}		
\usepackage{graphicx}
\usepackage{natbib}
\usepackage{doi}
\usepackage{algorithm}
\usepackage{algpseudocode}
\usepackage{amsmath}

\title{Error-Aware Curriculum Learning for  \\Biomedical Relation Classification}

\author{ {\hspace{1mm}Sinchani Chakraborty}\\
	Department of Computer Science and Engineering\\
	Indian Institute of Technology Kharagpur\\
	\texttt{sinchanichakraborty@gmail.com} \\
	\And
	{Sudeshna Sarkar} \\
	Department of Computer Science and Engineering\\
	Indian Institute of Technology Kharagpur\\
	\AND
	Pawan Goyal \\
        Department of Computer Science and Engineering\\
	Indian Institute of Technology Kharagpur\\
}

\renewcommand{\shorttitle}{\textit{arXiv} Template}

\hypersetup{
pdftitle={A template for the arxiv style},
pdfsubject={q-bio.NC, q-bio.QM},
pdfauthor={David S.~Hippocampus, Elias D.~Striatum},
pdfkeywords={First keyword, Second keyword, More},
}

\begin{document}
\maketitle

\begin{abstract}
Relation Classification (RC) in biomedical texts is essential for constructing knowledge graphs and enabling applications such as drug repurposing and clinical decision-making. We propose an error-aware teacher–student framework that improves RC through structured guidance from a large language model (GPT-4o). Prediction failures from a baseline student model are analyzed by the teacher to classify error types, assign difficulty scores, and generate targeted remediations, including sentence rewrites and suggestions for KG-based enrichment. These enriched annotations are used to train a first student model via instruction tuning. This model then annotates a broader dataset with difficulty scores and remediation-enhanced inputs. A second student is subsequently trained via curriculum learning on this dataset, ordered by difficulty, to promote robust and progressive learning. We also construct a heterogeneous biomedical knowledge graph from PubMed abstracts to support context-aware RC. Our approach achieves new state-of-the-art performance on 4 of 5 PPI datasets and the DDI dataset, while remaining competitive on ChemProt.
\end{abstract}


\section{Introduction}
Relation Extraction (RE) is a foundational task in Natural Language Processing that involves identifying and classifying semantic relationships between pairs of entities in textual data. When the task is reduced to classifying a relation type between two pre-identified entities, it is referred to as Relation Classification (RC). In the biomedical domain, it facilitates many downstream applications, including drug discovery, disease mechanisms, and the construction of biomedical knowledge graphs. 
Biomedical texts are often replete with long, compound sentences containing multiple nested entities and subtle semantic cues. For instance, a sentence may mention various drugs and diseases, with interactions that are negated, conditional, or context-dependent. Correctly classifying relationships in such settings requires an excellent syntactic understanding.  Extracting correct relations from such texts is challenging not only due to this linguistic complexity but also because many relations are context-sensitive, requiring background knowledge not explicitly stated in the text.

While supervised, classification-based methods have been the mainstream approach for RE in biomedical NLP, recent developments in Large Language Models (LLMs) have introduced prompt-based approaches that aim to address generalisation challenges via instruction tuning and in-context learning. However, despite their impressive capabilities, the closed-source nature and high per-token inference cost of many large models limit their accessibility and broader adoption. Additionally, empirical evidence shows that LLMs---even those with hundreds of billions of parameters like GPT-3.5 \cite{radford2018improving} or GPT-4 \cite{achiam2023gpt} through few-shot learning struggle to outperform smaller domain-specific models (e.g., PubMedBERT \cite{gu2021domain}) on biomedical information extraction tasks. On the other hand, standard fine-tuning is often brittle and sensitive to annotation noise and class imbalance.  

A promising direction involves the development of detailed error taxonomies \cite{andrade2024explaining, bassignana2024s, bassignana2022you} that categorise the types of mistakes classifiers make under specific training paradigms. This line of work has been instrumental in both general and biomedical domains, helping to pinpoint error-inducing instances in the training data that hinder robust generalisation.  In parallel, another line of work explores ways to augment training data \cite{jiang2024relation, hu2023gda, xu2016improved, guo2024few, zhao2025and} using outputs from larger, more powerful LLMs. This is especially common when budget constraints make direct use of large models impractical, so this strategy allows smaller models to benefit from the broader contextual knowledge encoded in larger models. 

Another set of strategies focuses on generating generic reasoning chains for enhancing task understanding \cite{ma2023chain, mustafa2025can, chen2024evaluating}. However, these approaches typically treat explanations as static reasoning chains and rarely incorporate a dynamic error analysis process that involves targeted remediation—a key ingredient in effective human teaching \cite{narciss2025learning}. Recent work has explored various techniques to enhance the learning dynamics of smaller models, especially when paired with stronger LLMs \cite{ying2024llms, he2025adaptmi, tan2024small, zhang2024small}. More recent frameworks, such as solution-guided fine-tuning, have shown promise in improving the learning dynamics of student models, particularly when guided by stronger LLMs \cite{bi2024enhancing}. These approaches demonstrate that exposing a student model to the reasoning behind correct answers can enhance both understanding and generalisation. Still, most of these techniques do not explicitly address the root causes of failure, relying instead on generic rationales that are not tailored to the model’s specific weaknesses.

In this work, we propose an error-driven student–teacher learning framework to guide small language models using targeted interventions from a stronger teacher model (GPT-4o \cite{gpt4o}). Our approach is motivated by prior studies on error taxonomies, particularly in biomedical relation classification (RC) \cite{bassignana2022you, bassignana2024s, andrade2024explaining}. We selectively integrate external knowledge from curated biomedical knowledge graph (BiomedKG), but only for examples where such augmentation addresses specific knowledge deficiencies, avoiding noisy, indiscriminate enrichment. The training begins with supervised fine-tuning of a student model \cite{gu2021domain} on labelled biomedical RC data. After this initial phase, we identify failure cases by ranking instances based on log probabilities, flagging those with the highest uncertainty. These difficult samples are then passed to the teacher model \cite{gpt4o}, which performs qualitative error analysis using an error taxonomy that includes categories such as negation handling, semantic ambiguity, and lack of world knowledge.

Each instance is annotated with a difficulty score (0–5) based on the number of distinct error types present. The teacher then provides tailored remediation strategies. For negation-related issues, the sentence is reformulated into clearer, simplified versions. Ambiguous instances are rewritten or enriched with contextual information from a biomedical knowledge graph. Sentences with multiple or long-distance entity dependencies are decomposed or restructured to improve clarity. Additionally, amplifiers and modals are explicitly highlighted when they alter the semantics of the entity pair. Crucially, the teacher also generates solution guidance—a structured, step-wise reasoning strategy that outlines how to approach the problem without giving away the final answer. Unlike standard Chain-of-Thought prompting, which often reveals the conclusion, we use an existing approach that isolates reasoning from resolution \cite{bi2024enhancing}, allowing the student to internalise task-specific decision strategies more effectively.

The student model is then fine-tuned in a second stage using an enriched training set that includes:
(i) the original supervised examples,
(ii) remediated and reformulated error samples,
(iii) solution-guided instances, and
(iv) knowledge graph-enriched examples when applicable. Due to the cost constraints of large model inference, the teacher intervenes only on a selected subset, and the student \cite{me-llama} is expected to mimic its behaviour. Teacher outputs are used for few-shot supervision, embedding both the rationale and correction into the new training data.

To maximise learning efficiency, we adopt a curriculum learning strategy \cite{cirik2016visualizing} in stage 2 of the second training stage. The second student model \cite{gu2021domain} is presented with examples in ascending order of difficulty: beginning with those correctly predicted in the initial fine-tuning, followed by progressively harder samples based on their error scores. This scaffolding enables the model to develop robust representations and gradually generalise to complex cases.

\section{Methodology}
\begin{figure}[t]
\centering
\includegraphics[width=0.9\columnwidth]{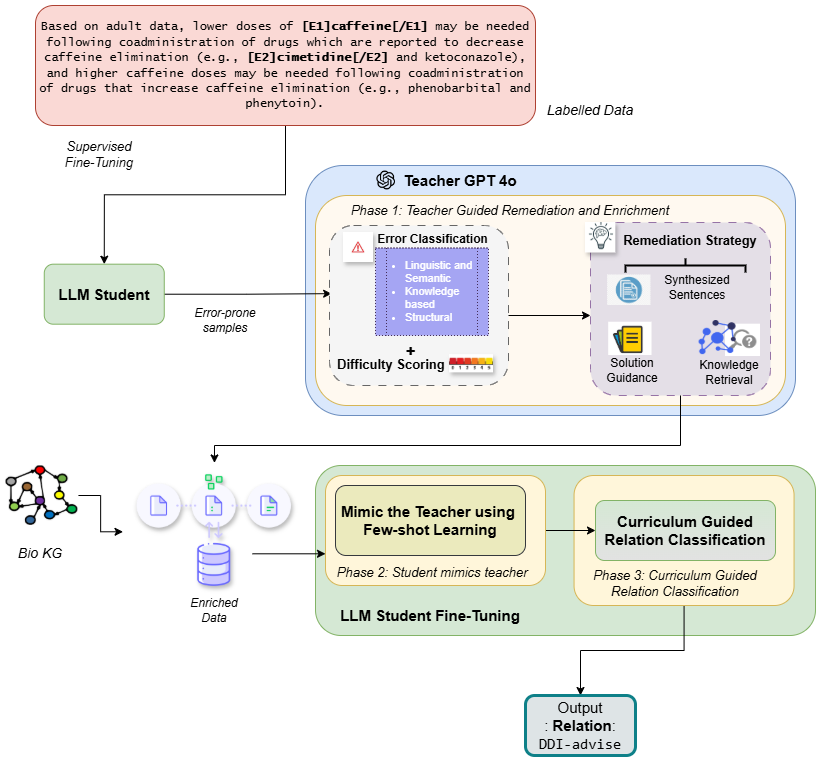}
    \caption{Teacher-student framework for error-aware relation classification with curriculum learning. The teacher GPT-4o guides the student by classifying errors, assigning difficulty scores, and generating remediation strategies. The student mimics the teacher using few-shot learning and is then trained with curriculum-guided relation classification.}
    \label{fig:student-teacher-framework}
\end{figure}

The framework depicted in the Figure \ref{fig:student-teacher-framework} shows our proposed methodology of Error-Aware Reasoning in a Teacher-Student Framework with Curriculum Learning. We explain the steps as follows:

\subsection*{Error-Aware Reasoning in a Teacher-Student Framework with Curriculum Learning }

As elaborated in the above steps, the teacher is a stronger LLM (GPT-4o) while the student is a smaller LLM (models with fewer than 10 billion parameters). The teacher module only intervenes in the error cases that the student model fails during a supervised fine-tuning setup.  The correct examples are left as is and are not intervened for the subsequent steps.  The framework, consisting of the Teacher module, encompasses the following: error-prone sample selection, scoring the hardness of the sample, classification of errors, provision of remediation strategies, and generation of solution guidance. The teacher module only works on a minimal set of data that provides scaffolding to the student \cite{nagesh2018exploration}. The student then learns to pick up better reasoning abilities and remediation techniques for errors. Also, this helps in managing the budget constraints of models that come with an inherent computational cost.
The student model then mimics the teacher model after getting fine-tuned on the generated metadata from the teacher. It generates all the remediation steps, solution guidance, error classification and hardness score for selected data samples. Finally, another student model is then trained using curriculum learning, implementing all the recommended changes and augmenting all the synthetic data. Algorithm 1 outlines the entire process.

\begin{algorithm}[t]
\caption{Error-Aware Curriculum Framework for Biomedical Relation Classification}
\label{alg:error-aware-curriculum}
\begin{algorithmic}[1]
\State \textbf{Input:} Dataset $\mathcal{D}$; Biomedical KG $\mathcal{K}$
\State \textbf{Parameters:} Teacher model $M_{\mathrm{Teacher}}$; Student models $M_{\mathrm{Student}}^{(1)}$, $M_{\mathrm{Student}}^{(2)}$; Threshold $\tau$
\State \textbf{Output:} Trained student models $M_{\mathrm{Student}}^{(1)}$, $M_{\mathrm{Student}}^{(2)}$

\State \textbf{Phase 1: Teacher-Guided Remediation and Enrichment}
\State Identify error instances: $\mathcal{D}_{\mathrm{error}} \gets \{x \in \mathcal{D} \mid \mathcal{L}(x) > \tau\}$
\ForAll{$(s_i, (e_1, e_2), R_{\mathrm{ref}}) \in \mathcal{D}_{\mathrm{error}}$}
    \State Predict error types $\epsilon_i$ from $s_i$
    \State Compute difficulty score $h_i$ from $\epsilon_i$
    \State Generate rewritten sentence $s'_i$, solution steps $S_i$, and error tag
    \If{KG support is required based on $\epsilon_i$}
        \State Retrieve top-$k$ triples $\mathcal{F}_i$ from $\mathcal{K}$
    \Else
        \State $\mathcal{F}_i \gets \emptyset$
    \EndIf
    \If{instance is ambiguous}
        \State Discard instance
    \ElsIf{new sentence is required}
        \State $\tilde{s}_i \gets \texttt{concat}(s'_i, S_i)$
    \Else
        \State $\tilde{s}_i \gets \texttt{concat}(s_i, S_i, \mathcal{F}_i)$
    \EndIf
\EndFor
\State Filter $\tilde{s}_i$ where $M_{\mathrm{Teacher}}(\tilde{s}_i) = R_{\mathrm{ref}}$ to form $\mathcal{D}_{\mathrm{rem}}$

\State \textbf{Phase 2: Mimicry by Student Model $M_{\mathrm{Student}}^{(1)}$}
\State $\mathcal{D}_{\mathrm{mimic}} \gets \mathcal{D}_{\mathrm{rem}} \cup \mathcal{D}_{\text{correct}}$
\State Train $M_{\mathrm{Student}}^{(1)}$ to jointly predict: error types, difficulty $h_i$, remediation, KG support
\State Generate augmented dataset $\mathcal{D}_{\mathrm{aug}} = \{(\hat{s}_i, (e_1, e_2), R_{\mathrm{ref}}, h_i)\}$

\State \textbf{Phase 3: Curriculum-Guided Training of $M_{\mathrm{Student}}^{(2)}$}
\State Partition $\mathcal{D}_{\mathrm{aug}}$ into buckets $B_1, \dots, B_5$ based on $h_i$
\For{$k = 1$ to $5$}
    \State $\mathcal{B} \gets \bigcup_{j=1}^{k} B_j$
    \State Train $M_{\mathrm{Student}}^{(2)}$ on $\mathcal{B}$ using multi-label loss
\EndFor
\State Compute total loss across curriculum stages

\State \textbf{Return} $M_{\mathrm{Student}}^{(1)}$, $M_{\mathrm{Student}}^{(2)}$
\end{algorithmic}
\end{algorithm}

We now describe the key modules of our proposed framework. Section~\ref{sec:teacher_strategy} details the dataset enrichment process carried out by the Teacher, while Section~\ref{sec:student_ch5} outlines the fine-tuning of the Student model using guided curriculum learning for relation classification.

\section{LLM as Teacher}
\label{sec:teacher_strategy}
In the Teacher-based strategy, the teacher is employed in two stages: first, in a zero-shot setting it performs structured error classification and assigns difficulty score to error samples to systematically identify and understand the causes of misclassifications (see ~\ref{subsec:Error}); and second, in a zero-shot setting, it provides targeted remediation strategies for refining the student (see ~\ref{subsec:Remediation}). Both of these are crucial for designing informed learning interventions and remediation from the teacher model.

\subparagraph{Preliminary}
We formalise a set of definitions and notations used throughout the remainder of the paper. The dataset $D$ is defined as follows:
\[
D = \left\{ (s, (e_1, e_2), R_{\mathrm{ref}}) \right\},
\]
where $s$ is the sentence whose relation has to be classified, $(e_1, e_2)$ are the pair of entities occurring in the sentence, among which the relation has to be classified, and $R_{\mathrm{ref}}$ is the reference label set of relations within which the classification has to be done. Given an input sentence $(s, (e_1, e_2))$, let $\hat{R}$ define the set of relations that are predicted by the target model. We define a relation label as incorrect when $\hat{R} \neq R_{\mathrm{ref}}$, which indicates either a missing relation or a prediction of an incorrect relation. On the other hand, a \textit{correct case} is defined as one where the prediction exactly matches the reference:
\[
\hat{R} = R_{\mathrm{ref}}.
\]

\subsection{Data Selection for Teacher Model using Log-Probability Thresholds}
\label{subsec:SFT}

We begin by performing \textit{supervised fine-tuning} of a smaller language model $\mathcal{M}_{\mathrm{student}}$ on a task-specific relation classification dataset:
\[
\mathcal{D} = \left\{ \left( s_i, (e_{1}^{(i)}, e_{2}^{(i)}), R_{\mathrm{ref}}^{(i)} \right) \right\}_{i=1}^{N},
\]
where $s_i$ is a sentence, $(e_{1}^{(i)}, e_{2}^{(i)})$ is a pair of entities appearing in the sentence, and $R_{\mathrm{ref}}^{(i)} \in \mathcal{R}$ is the relation expressed between them. The student model $\mathcal{M}_{\mathrm{student}}$, which may be an encoder-based model like PubMedBERT~\cite{pubmedbert} or a decoder-based architecture like Med-LLaMA~\cite{me-llama} with parameter-efficient adaptation, e.g., LISA~\cite{pan2024lisa}, is trained to predict a relation label from the candidate label space $\mathcal{R}$.

During training, we apply a multi-label \textit{binary cross-entropy loss} over all candidate labels to encourage accurate scoring across the label space. The loss for a single example is given by:
\[
\ell = - \sum_{r \in \mathcal{R}} \left[ y_r \log(\hat{y}_r) + (1 - y_r) \log(1 - \hat{y}_r) \right],
\]
where \( y_r \in \{0, 1\} \) indicates whether relation \( r \) belongs to the ground-truth set \( R_{\mathrm{ref}} \), and \( \hat{y}_r \in [0, 1] \) is the predicted probability for relation \( r \).

During training, we monitor the per-sample training loss \( \ell_i \) for each instance. These losses are used to identify samples that the student model finds difficult to learn. We compute the per-sample loss \( \ell(s, (e_1, e_2), R_{\mathrm{ref}}) \) and define a threshold \( \tau \) to isolate hard examples that are error-prone:
\[
\ell(s, (e_1, e_2), R_{\mathrm{ref}}) > \tau.
\]

Examples satisfying this condition are deemed error-prone and selected into a curated subset:
\[
\mathcal{D}_{\mathrm{error}} = \left\{ (s, (e_1, e_2), R_{\mathrm{ref}}) \in \mathcal{D}_{\mathrm{eval}} \mid \ell(s, (e_1, e_2), R_{\mathrm{ref}}) > \tau \right\}.
\]

Instead of treating all errors equally, we pass \( \mathcal{D}_\mathrm{error} \) to a stronger \textit{teacher model} \( \mathcal{M}_\mathrm{teacher} \), like GPT-4o \cite{gpt4o} which provides enriched outputs---such as corrected labels, rationales, or instructional steps---to guide refinement of the student model. The framework flexibly identifies challenging instances and directs additional supervision where it is most needed.

\begin{table*}[t]
\centering
\caption{Categorization and Description of Error Types}
\label{tab:error_taxonomy}
\renewcommand{\arraystretch}{1.2}
\begin{tabular}{p{3.5cm}|p{11cm}}  
\hline
\textbf{Category} & \textbf{Error Types and Descriptions} \\ 
\hline
Linguistic \& Semantic &
\begin{tabular}[t]{@{}l@{}}
\textbf{Negation:} Words that reverse meaning. \\
\textbf{Contrast:} Oppositional structures mislead models. \\
\textbf{Amplification/Modality:} Modifiers of intensity or certainty.
\end{tabular}
\\
\hline
Knowledge-based &
\textbf{Lack of Domain Knowledge:} Requires biomedical context. \\
\hline
Structural &
\begin{tabular}[t]{@{}l@{}}
\textbf{Multiple Entities:} Causes ambiguity in relation scope. \\
\textbf{Distant Entities:} Widely separated mentions.
\end{tabular}
\\
\hline
\end{tabular}
\end{table*}


\subsection{Error Classification and Difficulty Scoring}
\label{subsec:Error}
We selectively provide the student model's $\mathcal{M}_{\mathrm{student}}$ prediction failures $\mathcal{D}_{\mathrm{error}}$—those falling below a certain threshold (as detailed in Sub-section~\ref{subsec:SFT})—to a teacher model, GPT-4o. The error classification and difficulty-scoring steps are defined below.

\paragraph{Error Classification:}
We prompt a \textit{teacher model} to annotate each misclassified training instance by identifying the underlying types of linguistic or semantic errors. The error taxonomy used in our framework builds on prior work, which includes the following error types: \textit{linguistic and semantic}, \textit{knowledge-based}, \textit{structural}.
We adopt linguistically grounded categories such as negation, ambiguity, modality and insufficient knowledge from sentiment analysis \cite{andrade2024explaining}, which are broadly task-agnostic and capture fundamental linguistic and semantic challenges. We further adapt task-specific error types from relation classification and biomedical information extraction \cite{bassignana2022you,bassignana2024s,yadav2019feature}.

We hypothesise that targeting such errors and getting help for managing such instances from a capable teacher model can alleviate model errors and enhance student model performance. We use GPT-4o as our teacher model in a zero-shot prompting format to diagnose the misclassification of the error types discussed in Table\ref{tab:error_taxonomy}. Additionally, every error category is assigned a unique tag which the model uses to label misclassified sentences.

\paragraph{Difficulty scoring:}
\label{subsec:DifficultyScoring}
Each sentence in the \( \mathcal{D}_\text{error} \) may be annotated with zero or multiple error types to determine the estimation of solution confidence and facilitate subsequent task-specific classification. Primarily, the motivation arises from the observation that multiple interacting factors identified during error classification can contribute to sample difficulty. In particular, the presence of complex linguistic phenomena such as negation, contrast, or modality, as well as the requirement of external domain-specific knowledge, often corresponds to harder examples that challenge standard contextualised models. Based on these, we assign a difficulty score to samples. A hardness score $h: \mathcal{D}_\text{error} \rightarrow \{0, 1, 2, 3, 4, 5\}$ is computed based on the number of distinct error types:
\begin{itemize}
    \item $h = 0$: No error type detected
    \item $h = 1$ or $2$: One error type
    \item $h = 3$: Two error types
    \item $h = 4$ or $5$: Three or more error types
\end{itemize}
All error types are equally weighted. The type and number of remediations required by more difficult instances will be greater compared to easier ones. These scores facilitate subsequent curriculum-driven retraining by gradually introducing challenging examples.

\subsection{Remediation Strategies}
\label{subsec:Remediation}
In the second stage of our framework, we re-engage the teacher model \( \mathcal{M}_\mathrm{teacher} \) for \textit{active remediation} of errors identified in the dataset
\[
\mathcal{D}_\mathrm{error} = \{(x_i, y_i, \epsilon_i)\}_{i=1}^N,
\]
where each sentence-label pair \( (x_i, y_i) \in \mathcal{D} \subseteq (\mathcal{D}_\mathrm{error}) \), and \( \epsilon_i \in \mathcal{E} \) is the corresponding error category from a predefined taxonomy (Table~\ref{tab:error_taxonomy}). Here, \( x_i \) is the input sentence and \( y_i \) is the gold relation label between a given entity pair.

Unlike the previous stage, where it solely acts as an evaluator, here it acts as an informed tutor, providing corrective remediations for each error category and reasoning strategies for all error categories. The remediation strategy does not provide a blanket solution to all instances. To guide this behaviour, we provide a \textit{unified prompt} that includes global remediation rules \( \phi(\epsilon) \) for each error type \( \epsilon \in \mathcal{E} \). To ensure high-quality, interpretable guidance, we provide the teacher model with a set of rules and heuristics on how to handle each error category. It suggests the selective use of external knowledge from a biomedical knowledge graph (KG) we have constructed from a curated PubMed corpus (BiomedKG). This KG is queried only when the teacher model determines that sentence-level information alone is insufficient to resolve an error. For the reasoning strategy, we use solution guidance (\cite{bi2024enhancing}). These remediation solutions can be heuristic-driven, but the generation of solution guidance for every example is solely dependent on the model.  The solution guidance is different from conventional Chain-of-thought prompting and essentially consists of steps to reach the answer rather than answering on the spot.

\paragraph{Formal Remediation Specification.}
For each error instance in \( \mathcal{D}_\mathrm{error} \), \( \mathcal{M}_\text{teacher} \) generates a remediation tuple:

\[
(s_i, s'_i, S_i, \texttt{tag}_i),
\]
where:
\begin{itemize}
    \item \(s_i\): Original sentence.
    \item \( s'_i \): a simplified or restructured version of \( s_i \), when applicable,
    \item \( S_i \): stepwise solution guidance aiding the student model’s reasoning,
    \item \( \texttt{tag}_i \): Token(s) corresponding to the error class (e.g., \texttt{[\#\#\#KGLOOKUP]}, \texttt{[\#\#\#CON]}.
\end{itemize}

The remediation rules defined per error type are as follows:

\begin{itemize}
    \item \textbf{Negation Errors} \\ 
    Simplify the sentence and highlight negation explicitly. Annotate the negation cue with \texttt{[\#\#\#NEW\_NEG]}.

    \item \textbf{Ambiguity}  \\
    Signal semantic ambiguity by appending \texttt{[\#\#\#CON]}.Append \texttt{[\#\#\#CON\_KGLOOKUP]} to indicate the need for external knowledge, but do not synthesise new examples.

    \item \textbf{Amplification and Modality} \\
    Highlight amplifiers and modal verbs (e.g., ``strongly binds'', ``may'', ``could'') only when they influence \( (e_1, e_2) \). Append \texttt{[\#\#\#AMP]}

    \item \textbf{Knowledge Deficiency} \\
    Append \texttt{[\#\#\#KGLOOKUP]} to indicate the need for external knowledge. No new sentence is generated.

    \item \textbf{Multiple Entities} \\
    Decompose \( x_i \) into simpler sentences, each focused on a single entity pair. Annotate the original with \texttt{[\#\#\#NEW\_MULTI]}.

    \item \textbf{Long-Distance Dependencies}  \\
    If \( \text{dist}(e_1, e_2) > \delta \) (e.g., \( \delta = 20 \)), restructure the sentence. Annotate with \texttt{[\#\#\#NEW\_DIST]}.
\end{itemize}

\paragraph{Solution Guidance Specification.}
The solution guidance component \( S_i \) provides structured, interpretable reasoning steps for each instance \( (x_i, y_i, \epsilon_i) \), guiding the model toward the correct inference pathway without revealing the answer. Inspired by \cite{bi2024enhancing}, this approach diverges from conventional Chain-of-Thought \cite{wei2022chain} prompting by emphasising instructional decomposition over answer generation. It scaffolds reasoning through intermediate subgoals, encouraging the student model to internalise error-specific heuristics. By decoupling guidance from prediction, \( S_i \) facilitates more deliberate, pedagogically grounded fine-tuning, enhancing the model’s capacity for robust, generalizable relational understanding.

This format enables interpretable, traceable generation and better instructional grounding during student model training.
Following the synthesis step, the original error-prone samples \( D_{\mathrm{error}} \) are replaced with the newly generated versions wherever remediation was applied. For cases where remediation requires consulting the biomedical knowledge graph, relevant triples are appended. These triples are retrieved using an embedding-based top-\(k\) similarity search, following the method described by \cite{yao2025exploring}. The retrieved KG facts are mentioned in the prompt -- Remediation Prompt Format, along with the modified sentence, resulting in a semantically enriched input.

\paragraph{Data Cleaning:}
\label{para:dataclean}
We begin by augmenting the dataset with newly synthesized sentences. Each input sentence is then revised by appending the corresponding generated solution guidance. Next, we scan for the tags \texttt{[\#\#\#CON\_KGLOOKUP]} and \texttt{[\#\#\#KGLOOKUP]} to extract relevant triples from an external knowledge graph. These revised sentences are subsequently passed to the teacher model, which performs relation classification. The predicted relation label is compared against the ground truth label. Only those instances for which the teacher’s prediction matches the reference label are retained. Formally, the remediated dataset is filtered as follows:

\[
\mathcal{D}_{\mathrm{rem}} = \left\{ (\tilde{s}_i, (e_1, e_2), R_{\mathrm{ref}}) \mid M_T(\tilde{s}_i) = R_{\mathrm{ref}} \right\}
\]

The resulting filtered instances serve as refined few-shot examples for training the student model.

\textbf{Integrating BiomedKG triples:} Following the approach of \cite{yao2025exploring}, we sample \( K = 5 \) neighbouring entities from BiomedKG for each target entity, explicitly excluding the entity itself. These neighbours are then used to construct context-rich prompts that provide auxiliary knowledge to guide the student model during training.

\section{Student Model Finetuning}
\label{sec:student_ch5}

To transfer the capabilities of the teacher model to a smaller student model, we adopt a two-stage fine-tuning strategy. Specifically, it decomposes the overall process into two stages : (i) Mimicking the Teacher via Few-Shot Learning, (ii) Relation Classification using solution guidance. Both stages use the LISA fine-tuning method, which helps in resource-constrained optimisation. 

\subsection{Stage 1: Mimicking the Teacher via Few-Shot Learning}
\label{subsec:Student1}

The student model is expected to act like the teacher and is not expected to directly perform the relation classification task. Instead, it focuses on teaching the student model to replicate the instructional behaviour of the teacher model, as described in Sections~\ref{subsec:Error} and~\ref{subsec:Remediation}. Specifically, the student is trained to (i) identify error types given a sentence according to the error taxonomy Table \ref{tab:error_taxonomy}, (ii) assign relative difficulty scores between (0-5) to relation instances, (iii) suggest appropriate remediation heuristics (e.g., simplification, splitting, knowledge augmentation), and (iv) produce solution-guidance that assist the downstream inference.

For this stage, we curate a training set of 1000, 2000, 3000 instances referring to the initial supervised relation classification stage (see Section~\ref{subsec:SFT}). The training data used in the present case comprises of two components: (a) the remaining error-prone instances which are not the part of \( \mathcal{D}_\mathrm{error} \) that did not undergo intervention (described in Section ~\ref{subsec:Remediation}), (b) randomly sampled subset of correctly predicted examples from the same training distribution, included to preserve label diversity and ensure balanced learning. We represent the mimic dataset as follows:
\[
\mathcal{D}_{\mathrm{mimic}} = \mathcal{D}_{\mathrm{rem}} \cup \mathrm{Sample}(\mathcal{D}_{\mathrm{correct}})
\]

The student model is trained using a few-shot classification setup. The few-shot exemplars are the final retained, refined and augmented sentences corresponding to the error-prone instances present in \( \mathcal{D}_\mathrm{error} \) generated after remediation from the teacher model and left after data cleaning as discussed in module \ref{para:dataclean}. The exemplars comprise augmenting newly rewritten sentences and appending knowledge graph triples -- each introduced at appropriate points based on the teacher model recommendation, along with solution guidance for every exemplar. The few-shot examples ensure that the student model receives exposure to both difficult and canonical instances, allowing it to better generalise the classification behaviour exhibited by the teacher model.

\subsubsection{Efficient Instruction Tuning with LISA in a Teacher– Student Framework}

Recent work demonstrated that parameter-efficient tuning methods such as LoRA \cite{hu2021lora} primarily update the embedding and output layers of large language models (LLMs), leaving the intermediate layers underutilised. Building on this insight, they proposed \textbf{LISA (Layer Importance Sampling for Adaptation)}, which was introduced to selectively fine-tune only the most informative layers, significantly reducing GPU memory usage and training time while maintaining strong downstream performance \cite{pan2024lisa}.

In our \textit{teacher--student} framework, we leverage LISA to perform \textbf{instruction tuning on the MedLLaMA-3-8B model}, which acts as \textbf{($M_{\mathrm{Student}}^{(1)}$)} ($M_{\mathrm{Student}}^{(1)}$). The goal is to align MedLLaMA-3-8B with the output behaviour of a powerful \textbf{Teacher LLM (GPT-4o)}, which performs error-aware relation classification. Given the large number of parameters in the student model, full fine-tuning would be computationally expensive. LISA enables us to efficiently distil the teacher’s reasoning into a small, highly relevant subset of parameters.

LISA computes a layer-wise importance score \( I_l \) for each transformer layer \( l \in \{1, \ldots, L\} \), based on sensitivity metrics such as gradient norms or Fisher information. The top-\( k \) layers with the highest importance form a subset \( \mathcal{S} \subset \{1, \ldots, L\} \), which are selectively updated:
\[
\mathcal{S} = \mathrm{Top}_k(\{I_1, I_2, \ldots, I_L\})
\]

This approach enables instruction tuning to proceed at approximately \textbf{1.5$\times$ the speed of LoRA} on 7B-scale models, with competitive performance.

To ensure stable optimisation, LISA introduces a regularised objective function:
\[
f_{\mathrm{reg}}(\mathbf{w}) = f(\mathbf{w}) + \frac{1}{2} \mathbf{w}^\top S \mathbf{w}
\]
where \( \mathbf{w} \) denotes the trainable parameters in \( \mathcal{S} \), and \( S \) is a diagonal positive semi-definite matrix encoding per-layer regularisation weights. This term constrains model complexity and aids generalisation.

Moreover, LISA offers a convergence guarantee over \( T \) iterations:
\[
\frac{1}{T} \sum_{t=1}^{T} f_{\mathrm{reg}}(\mathbf{w}_t) - f^*_{\mathrm{reg}} \leq \mathcal{O}\left(\frac{1}{\sqrt{T}}\right)
\]

During instruction tuning, Student 1 ($M_{\mathrm{Student}}^{(1)}$) learns to replicate the teacher’s outputs, including \textit{error types}, \textit{difficulty scores}, and \textit{solution guidance}. This fine-tuned student then serves as a knowledge-enriched intermediary for the final relation classification model Student 2 ($M_{\mathrm{Student}}^{(2)}$), which is trained using curriculum learning strategies.

By integrating LISA into the instruction tuning phase, we enable scalable and compute-efficient adaptation of \textbf{MedLLaMA-3-8B} to domain-specific biomedical tasks. This facilitates high-quality knowledge distillation and downstream performance, even in resource-constrained training environments.

At the end of this phase, the student model can generate enriched annotations—including error types, remediation strategies, and knowledge suggestions—for all 2,000 training examples. Final annotated samples :
\[
\mathcal{D}_{\mathrm{aug}} = \{(\hat{s}_i, (e_1, e_2), R_{\mathrm{ref}}, h_i)\}
\]

\subsection{Stage 2: Curriculum-Guided Relation Classification}

\begin{figure}[t]
\centering
\includegraphics[width=0.9\columnwidth]{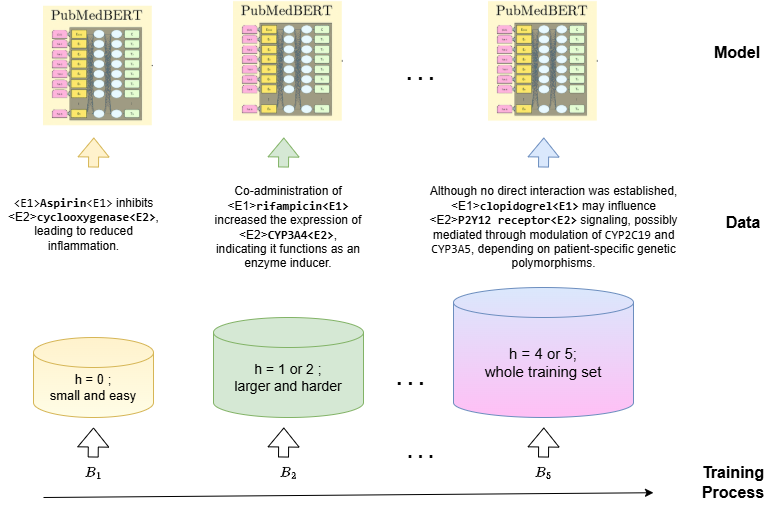}
\caption{Overview of the curriculum-guided training phase of the second student model, which performs multi-label biomedical relation classification. 
Following the Baby Steps strategy, training proceeds from easy to hard by cumulatively including buckets $\mathcal{B}_1$ to $\mathcal{B}_5$. This progressive approach enables the student model (PubMedBERT) to first learn from simpler instances and gradually generalise to complex ones.
}
\label{fig:CL_snapshot}
\end{figure}

In the subsequent stage of our student training pipeline, we use another smaller student language model that is specifically focused on performing multi-label relation classification. Unlike the first student model, which emulated the teacher's behaviour, this model is optimised to perform the task of Relation Classification. After fine-tuning the smaller model in the previous section ~\ref{subsec:Student1}, the data serves as input in this stage. The new synthesised sentences are augmented to the training data, replacing the error-causing ones wherever the student model suggested; those instances that required knowledge graph support are provided. We use the original format of relation classification as used for instruction tuning, plus the solution guidance and knowledge graph triples. 

This enriched dataset is passed to the second student model ($M_{\mathrm{Student}}^{(2)}$), which is fine-tuned using a curriculum learning approach \cite{bengio2009curriculum}, which performs the task of relation classification. The training curriculum is constructed based on heuristic difficulty scores suggested by the student model in section mimicking the teacher (see \ref{subsec:DifficultyScoring}). The training paradigm ensures that the model encounters simpler instances early and progressively advances to more complex examples. Each training instance was scored (0--5), as predicted by the previous student model during the error analysis and difficulty scoring phase, while it was learned by mimicking the teacher. These scores were based on the number of error types assigned; a higher count of error tags indicated a more difficult sample. An instance requiring multiple interventions (e.g., sentence rewriting + KG appending + solution guidance) is considered more difficult and assigned a higher score (e.g., 5), while canonical or lightly remediated instances receive lower scores (e.g., 1 or 2).
Based on the assigned difficulty scores, we divide the training dataset into $K = 5$ buckets, denoted as $\{\mathcal{B}_1, \mathcal{B}_2, \dots, \mathcal{B}_5\}$. Each bucket $\mathcal{B}_k$ contains instances that share the same level of difficulty—-with $\mathcal{B}_1$ holding the easiest examples (score 0) and $\mathcal{B}_5$ containing the most challenging ones (score 5) as shown in \ref{fig:CL_snapshot}. These scores are derived from factors such as linguistic complexity (e.g., negation, contrast), the need for external knowledge (e.g., KG triples), the number of remediation strategies applied, and the confidence of the teacher model in resolving the instance.

We use  \textbf{Baby Steps curriculum learning strategy} proposed by \cite{cirik2016visualizing}, where training starts with the easiest bucket $\mathcal{B}_1$. As training progresses, we incrementally add the next bucket $\mathcal{B}_2$, then $\mathcal{B}_3$, and so on, while retaining all previously seen data in the training pool as shown in Figure \ref{fig:CL_snapshot}. This cumulative approach helps the model gradually handle increasingly difficult examples while reinforcing earlier knowledge, thereby avoiding catastrophic forgetting.

At each stage $k$, the model is trained on the union of all buckets from $\mathcal{B}_1$ to $\mathcal{B}_k$, denoted as $\mathcal{B}_{1:k} = \bigcup_{j=1}^{k} \mathcal{B}_j$. The empirical risk (loss) at curriculum stage $k$ is calculated as:

\[
\mathcal{L}_k = \frac{1}{|\mathcal{B}_{1:k}|} \sum_{(x_i, y_i) \in \mathcal{B}_{1:k}} \ell_{\mathrm{multi}}(f_{\theta}(x_i), y_i)
\]

where $f_{\theta}$ is the model with parameters $\theta$.

We use the sigmoid binary cross-entropy as the multi-label classification loss:

\[
\ell_{\mathrm{multi}}(\hat{y}, y) = -\sum_{j=1}^{L} \left[ y_j \log \hat{y}_j + (1 - y_j) \log (1 - \hat{y}_j) \right]
\]

where $L$ is the number of relation classes, $y \in \{0,1\}^L$ is the ground-truth multi-label vector, and $\hat{y} = f_{\theta}(x)$ is the predicted output.

Finally, the overall curriculum-aware training loss is the sum of all stage-wise losses:

\[
\mathcal{L}_{\mathrm{total}} = \sum_{k=1}^{K} \mathcal{L}_k
\]

This training regime enables the model to learn from simpler instances first and gradually develop the capacity to classify more complex relation types, all while maintaining stable performance and preventing degradation in previously acquired knowledge.

This combined strategy of difficulty-aware sequencing and parameter-efficient fine-tuning enables the student model to develop deeper reasoning skills while remaining scalable on modest hardware. This design mimics human learning patterns and has been shown to improve training stability and generalisation.
\subsection{Results}

\subsubsection{Comparison with Existing Methods}

\paragraph{Comparative Analysis on PPI Datasets:}

Table \ref{tab:ppi_results} presents the F1-scores of various models across four standard Protein-Protein Interaction (PPI) benchmarks: AiMed, IEPA, HPRD50, and LLL. The comparative performance across five benchmark Protein–Protein Interaction (PPI) datasets demonstrates the effectiveness of our proposed \textit{Error-Aware Curriculum Learning} strategy. Our approach consistently achieves state-of-the-art results on \textbf{AiMed}, \textbf{IEPA}, \textbf{HPRD50}, and \textbf{LLL}, and performs competitively on \textbf{BioInfer}, slightly trailing only the best-performing Tree Transformer-based model~\cite{roy2023extracting}.

\begin{table*}[t]
\centering
\small 
\caption{Performance (F1 Scores) of various models on PPI datasets. Our proposed method is marked with an asterisk. The best score is bold faced.}
\label{tab:ppi_results}
\begin{tabular}{|l|c|c|c|c|c|}
\hline
\textbf{Method} & \textbf{AiMed} & \textbf{BioInfer} & \textbf{IEPA} & \textbf{HPRD50} & \textbf{LLL} \\
\hline
Error-Aware Curriculum Learning (Ours) & \textbf{95.6*} & 96.72* & \textbf{95.90*} & \textbf{95.06*} & \textbf{94.90*} \\
\hline
Tree Transformers + Heterogeneous GAT~\cite{roy2023extracting} & 94.66 & \textbf{97.81} & 93.47 & 94.01 & 94.14 \\
\hline
BioBERT + Relation Context~\cite{park2022extracting} & 92.0 & 91.3 & 87.4 & 88.2 & 89.4 \\
\hline
Tree Transformers + Heterogeneous GAT~\cite{roy2023identifying} & 91.23 & 96.97 & 87.28 & 93.11 & 93.52 \\
\hline
Graph-based~\cite{fei2021span} & 88.27 & 96.21 & 83.90 & 89.57 & 92.86 \\
\hline
PubMedBERT\_SLL\_Att~\cite{su2022investigation} & 83.1 & -- & -- & -- & -- \\
\hline
BioBERT + ADVBERT-LSTM/CNN~\cite{tang2022protein} & 83.93 & -- & 84.88 & 84.78 & 88.65 \\
\hline
BioBERT + CLEK~\cite{su2021biobert_cleck} & 82.70 & -- & -- & -- & -- \\
\hline
BioBERT~\cite{lee2019biobert} & 81.0 & 78.81 & 78.81 & -- & 86.84 \\
\hline
PubMedBERT~\cite{gu2021domain} & 82.1 & 78.49 & 78.49 & 65.0 & 85.42 \\
\hline
GPT-4~\cite{rehana2024evaluating} & -- & -- & 71.54 & 78.73 & 86.49 \\
\hline
Att-sdpLSTM~\cite{yadav2019feature} & 93.29 & 81.68 & 76.25 & 78.73 & 83.92 \\
\hline
\end{tabular}
\end{table*}

On the \textit{AiMed} dataset, our method attains an F1 score of \textbf{95.6}, outperforming all existing baselines, including Tree Transformers + Heterogeneous GAT (94.66) and BERT + Relation Context (92.0). This highlights our model's ability to generalize from limited training data and manage the high linguistic variability in AiMed.

In the \textit{BioInfer} corpus, although Tree Transformers + Hetero. GAT achieves the highest F1 score (97.81), our model delivers a very competitive score of \textbf{96.72}, surpassing most other transformer-based and graph-based methods. This indicates that our curriculum design effectively captures complex semantic patterns even in large-scale and well-annotated corpora.

For \textit{IEPA}, our method leads with \textbf{95.90}, improving significantly over Tree Transformer-based methods (93.47 and 87.28) and BioBERT variants (as low as 78.81). Similarly, on \textit{HPRD50}, our model reaches \textbf{95.06}, outperforming the next best method (94.01) and showing robustness in small datasets with fine-grained protein annotations.

Finally, on \textit{LLL}, our model achieves \textbf{94.90}, marginally better than the Tree Transformer baseline (94.14) and clearly ahead of traditional transformer and LSTM-based approaches, which range between 83–89.

Overall, this consistent superiority across datasets of varying sizes and characteristics demonstrates the benefit of \textit{error-aware feedback loops} and \textit{difficulty-guided learning progression} in enhancing relational reasoning in biomedical texts. Notably, traditional fine-tuning approaches using BioBERT or PubMedBERT underperform in comparison, reinforcing the value of structured guidance in training deep models for complex biomedical NLP tasks.

\paragraph{Comparative Analysis on DDI:}

Table \ref{tab:ddi2013_results} summarises model performance on the DDI dataset. The proposed Error-Aware Curriculum Learning framework achieves a new state-of-the-art F1 score of 96.30, outperforming all previous models. This substantial improvement highlights the effectiveness of integrating error-based remediation and difficulty-aware learning into the training loop. By prioritizing harder examples and correcting specific model weaknesses, this method achieves more robust generalization across diverse biomedical sentence structures.

Among previous leading models, the Tree-structured + Heterogeneous Graph approach by Roy et al. (2023)\cite{roy2023extracting} yields the second-best performance with an F1 score of 95.20. This method effectively combines syntactic tree information with heterogeneous graph representations to encode structural and semantic dependencies, thereby enhancing relational inference. Similarly, the Graph-based model introduced by Fei et al.(2021)\cite{fei2021span} reports an F1 score of 93.40, demonstrating the efficacy of graph neural networks in modeling contextual relationships and hierarchical interactions between biomedical entities.

Models that explicitly incorporate domain knowledge show moderately strong performance. For example, the HKG-DDI model by Asada et al.(2023)\cite{asada2023integrating}, which leverages heterogeneous knowledge graphs, attains an F1 score of 85.40, reflecting the benefits of external knowledge integration. The DDMS-CNN model\cite{asada2021using}, based on multi-scale convolutional networks, achieves 84.08, indicating that local contextual features alone are insufficient to fully capture the complex patterns present in DDI contexts.

In the context of large language models (LLMs), LLaMA2-13B and MedLLaMA 13B achieve F1 scores of 91.73 and 91.33, respectively~\cite{touvron2023llama,xie2024me}. These results indicate that general-purpose LLMs, even when adapted to biomedical domains, still lag behind task-specific models that are structurally informed and error-aware. While their generalization ability is notable, their lack of task-level supervision and fine-grained learning objectives limits their effectiveness in nuanced biomedical extraction tasks.

Biomedical domain-pretrained language models, such as SciBERT\cite{beltagy2019scibert},   BioLinkBERT\cite{yasunaga2022linkbert}, BioBERT\cite{lee2019biobert}, and PubMedBERT\cite{gu2021domain}, achieve F1 scores in the range of 79.00–83.35. Although domain-specific pretraining enhances their baseline performance, these models often lack mechanisms for relational reasoning and adaptive supervision. Additionally, GPT-4~\cite{achiam2023gpt} yields an F1 score of 77.00 in a zero-shot setting, further underscoring the limitations of general-purpose prompting for highly specialized biomedical tasks.

In summary, this comparative evaluation reveals that structural modeling, knowledge integration, and difficulty-aware learning significantly boost performance on the DDI 2013 dataset. The proposed Error-Aware Curriculum Learning framework leverages all three components and thus establishes a new performance upper bound, demonstrating its effectiveness in advancing biomedical relation extraction.

\begin{table}[t]
\centering
\renewcommand{\arraystretch}{1.2}
\small
\begin{tabular}{|p{7cm}|c|}
\hline
\multicolumn{2}{|c|}{\textbf{DDI Dataset}} \\
\hline
\textbf{Method} & \textbf{F1 Score} \\
\hline
Error-Aware Curriculum Learning (Ours) & \textbf{96.30*} \\
Tree-structured + Heterogeneous Graph~\cite{roy2023extracting} & 95.20 \\
Graph-based~\cite{fei2021span} & 93.40 \\
HKG-DDI~\cite{asada2023integrating} & 85.40 \\
DDMS-CNN~\cite{asada2021using} & 84.08 \\
LLaMA2-13B~\cite{touvron2023llama} & 91.73 \\
LLaMA2-7B~\cite{touvron2023llama} & 89.44 \\
MedLLaMA 13B~\cite{xie2024me} & 91.33 \\
SciBERT~\cite{beltagy2019scibert} & 81.09 \\
BioLinkBERT~\cite{yasunaga2022linkbert} & 83.35 \\
BioBERT~\cite{lee2019biobert} & 79.00 \\
PubMedBERT~\cite{gu2021domain} & 82.36 \\
RHCNN~\cite{sun2019drug} & 75.48 \\
GPT-4~\cite{achiam2023gpt} & 77.00 \\
\hline
\end{tabular}
\caption{F1 scores of various methods on the DDI 2013 dataset. Our method is marked with an asterisk. The best score is bold faced.}
\label{tab:ddi2013_results}
\end{table}

\paragraph{Comparative Analysis on Chemprot:}

Table ~\ref{tab:ChemProt_Score} reports the F1-scores of various models evaluated on the ChemProt dataset.  The performance comparison on the ChemProt dataset (Table~\ref{tab:ChemProt_Score}) highlights the competitiveness of our proposed \textit{Error-Aware Curriculum Learning} approach, which achieves an F1-score of \textbf{88.91}, surpassing several recent large-scale and hybrid architectures. Although the highest performance is obtained by the prompt-based Bio-RoBERTa model~\cite{yeh2022decorate} (90.09), our method closely follows, outperforming other advanced models such as SciFive (88.95), LLaMA2-13B with BiomedRAG~\cite{li2025biomedrag} (88.83), and T5 slim decoder combined with BERT-GAT~\cite{kim2023biomedical} (87.46).

Notably, our method exceeds the performance of traditional transformer-based models like PubMedBERT (77.52) and BioBERT (76.46), as well as multi-modal and graph-based models such as ChemicalBERT + GCN (80.21) and BicapBERT (82.12). Furthermore, zero-shot variants of LLaMA2 and MedLLaMA13B show significantly weaker performance (77.48 and 50.52 respectively), emphasizing the importance of task-specific fine-tuning and guided learning strategies.

Overall, our model demonstrates strong generalization capabilities in relation classification tasks involving chemical–protein interactions, and its performance underscores the benefit of integrating error signals and difficulty-aware learning into biomedical NLP pipelines.

\begin{table}[t]
\centering
\small
\renewcommand{\arraystretch}{1.3}
\begin{tabular}{|p{6.9cm}|c|}
\hline
\multicolumn{2}{|c|}{\textbf{ChemProt Dataset}} \\
\hline
\textbf{Model Configuration} & \textbf{F1 Score} \\
\hline
Error-Aware Curriculum Learning (Ours) & 88.91* \\
Prompt-based Bio-RoBERTa-base~\cite{yeh2022decorate} & \textbf{90.09} \\
SciFive (PMC)~\cite{phan2021scifive} & 88.95 \\
LLaMA2-13B + BiomedRAG~\cite{li2025biomedrag} & 88.83 \\
T5 Slim Decoder + BERT-GAT~\cite{kim2023biomedical} & 87.46 \\
MolBERT~\cite{sanger2025knowledge} & 78.22 \\
BicapBERT~\cite{chen2024relational} & 82.12 \\
ChemicalBERT + GCN~\cite{qin2020chemical} & 80.21 \\
PubMedBERT~\cite{gu2021domain} & 77.52 \\
LLaMA2-13B (Zero-shot)~\cite{li2025biomedrag} & 77.48 \\
BioBERT~\cite{lee2019biobert} & 76.46 \\
MedLLaMA-13B (Zero-shot)~\cite{li2025biomedrag} & 50.52 \\
\hline
\end{tabular}
\caption{F1 scores on the ChemProt dataset for various models. Our proposed method is marked with an asterisk. The best score is highlighted in bold.}
\label{tab:ChemProt_Score}
\end{table}

\subsubsection{Ablation Study}

\paragraph{DDI classification performance across different model configurations.}

The results in the table \ref{tab:DDI_ablation_ch5} demonstrate that individually applying solution guidance, error remediation, or knowledge graph support yields moderate improvements over the PubMedBERT baseline. Among individual components, KG integration shows the highest boost (88.03 F1). However, combining error remediations with KG results in a significant gain (92.01 F1), and the full combination of remediation, KG, and solution guidance achieves the best performance at 96.30 F1. This highlights the complementary benefits of structured remediation, external knowledge, and instructional guidance in enhancing biomedical relation classification.

\begin{table}[t]
\centering
\renewcommand{\arraystretch}{1.3}
\small
\begin{tabular}{|p{6.3cm}|c|}
\hline
\textbf{Model Configuration} & \textbf{F1-score} \\
\hline
PubMedBERT (Baseline) & 81.90 \\
\hline
Error-Aware Curriculum Learning: \newline
\quad GPT-4o as $M_{\mathrm{Teacher}}$ \newline
\quad Bio-Medical-LLaMA-3-8B as $M_{\mathrm{Student}}^{(1)}$ \newline
\quad PubMedBERT as $M_{\mathrm{Student}}^{(2)}$ & -- \\
\hline
Only Solution Guidance & 83.70 \\
\hline
Only Error Remediation (No KG) & 82.65 \\
\hline
Only Knowledge Graph & 88.03 \\
\hline
Error Remediation + Knowledge Graph & 92.01 \\
\hline
Solution Guidance + Knowledge Graph & 90.40 \\
\hline
Solution Guidance + Error Remediation & 86.80 \\
\hline
All Combined (Remediation + KG + Solution Guidance) & \textbf{96.30} \\
\hline
\end{tabular}
\caption{
DDI classification performance (F1 scores) across different model configurations 
using the proposed Error-Aware Curriculum Learning framework.
}
\label{tab:DDI_ablation_ch5}
\end{table}

\paragraph{Effect of Curriculum Learning}

\begin{table}[ht]
\centering
\renewcommand{\arraystretch}{1.2}
\small
\begin{tabular}{|l|c|c|c|}
\hline
\textbf{Dataset} & \textbf{Without CL} & \textbf{With CL} & \textbf{Gain ($\Delta$ F1)} \\
\hline
\textbf{DDI}       & 94.12 & 96.30 & +2.18 \\
\textbf{ChemProt}  & 88.36 & 88.91 & +0.55 \\
\textbf{PPI (AiMed)} & 95.10 & 95.60 & +0.50 \\
\hline
\end{tabular}
\caption{
Effect of curriculum learning on biomedical relation classification using our proposed
Error-Aware Curriculum Learning framework. For PPI, we consider the AiMed dataset.
Curriculum-guided training yields marginal improvements across all datasets.
}
\label{tab:curri_ablation_ch5}
\end{table}

From Table \ref{tab:curri_ablation_ch5} we can see that the application of curriculum learning yields consistent but modest gains across all datasets. Compared to standard training, the F1 scores improved by +2.18 on DDI, +0.55 on ChemProt, and +0.50 on PPI. While the improvements are not large, they suggest that curriculum learning contributes to better generalisation and training stability in the student model.

\subsection{Conclusion}
We introduced a novel error-aware framework for biomedical relation classification that integrates structured remediation, curriculum learning, and external knowledge grounding. By leveraging a powerful teacher model to annotate training data with error types, solution guidance, and difficulty scores, our two-stage student learning pipeline enables targeted improvement. The use of knowledge graph triples and guided augmentation helps disambiguate complex biomedical contexts, while the LISA framework allows efficient fine-tuning on large models even with limited computational resources. Our results demonstrate that combining explanation-driven supervision with curriculum learning significantly enhances relation classification performance, particularly on challenging biomedical benchmarks.

\bibliographystyle{unsrtnat}
\bibliography{references}  

\begin{thebibliography}{54}
\providecommand{\natexlab}[1]{#1}
\providecommand{\url}[1]{\texttt{#1}}
\expandafter\ifx\csname urlstyle\endcsname\relax
  \providecommand{\doi}[1]{doi: #1}\else
  \providecommand{\doi}{doi: \begingroup \urlstyle{rm}\Url}\fi

\bibitem[Radford et~al.(2018)Radford, Narasimhan, Salimans, Sutskever, et~al.]{radford2018improving}
Alec Radford, Karthik Narasimhan, Tim Salimans, Ilya Sutskever, et~al.
\newblock Improving language understanding by generative pre-training.
\newblock \emph{OpenAI Blog}, 2018.

\bibitem[Achiam et~al.(2023)Achiam, Adler, Agarwal, Ahmad, Akkaya, Aleman, Almeida, Altenschmidt, Altman, Anadkat, et~al.]{achiam2023gpt}
Josh Achiam, Steven Adler, Sandhini Agarwal, Lama Ahmad, Ilge Akkaya, Florencia~Leoni Aleman, Diogo Almeida, Janko Altenschmidt, Sam Altman, Shyamal Anadkat, et~al.
\newblock Gpt-4 technical report.
\newblock \emph{arXiv preprint arXiv:2303.08774}, 2023.

\bibitem[Gu et~al.(2021)Gu, Tinn, Cheng, Lucas, Usuyama, Liu, Naumann, Gao, and Poon]{gu2021domain}
Yu~Gu, Robert Tinn, Hao Cheng, Michael Lucas, Naoto Usuyama, Xiaodong Liu, Tristan Naumann, Jianfeng Gao, and Hoifung Poon.
\newblock Domain-specific language model pretraining for biomedical natural language processing.
\newblock \emph{ACM Transactions on Computing for Healthcare (HEALTH)}, 3\penalty0 (1):\penalty0 1--23, 2021.

\bibitem[Andrade et~al.(2024)Andrade, Cunha, Fonseca, Pagano, Santos, Pagano, Rocha, and Gon{\c{c}}alves]{andrade2024explaining}
Claudio Andrade, Washington Cunha, Guilherme Fonseca, Ana Pagano, Luana Santos, Adriana Pagano, Leonardo Rocha, and Marcos Gon{\c{c}}alves.
\newblock Explaining the hardest errors of contextual embedding based classifiers.
\newblock In \emph{Proceedings of the 28th Conference on Computational Natural Language Learning}, pages 419--434, 2024.

\bibitem[Bassignana et~al.(2024)Bassignana, Van Der~Goot, and Plank]{bassignana2024s}
Elisa Bassignana, Rob Van Der~Goot, and Barbara Plank.
\newblock What’s wrong with your model? a quantitative analysis of relation classification.
\newblock In \emph{Proceedings of the 13th Joint Conference on Lexical and Computational Semantics (* SEM 2024)}, pages 252--263, 2024.

\bibitem[Bassignana and Plank(2022)]{bassignana2022you}
Elisa Bassignana and Barbara Plank.
\newblock What do you mean by relation extraction? a survey on datasets and study on scientific relation classification.
\newblock \emph{arXiv preprint arXiv:2204.13516}, 2022.

\bibitem[Jiang et~al.(2024)Jiang, Li, and Chen]{jiang2024relation}
Yizhi Jiang, Jinlong Li, and Huanhuan Chen.
\newblock Relation classification via bidirectional prompt learning with data augmentation by large language model.
\newblock In \emph{Proceedings of the 2024 Joint International Conference on Computational Linguistics, Language Resources and Evaluation (LREC-COLING 2024)}, pages 13885--13897, 2024.

\bibitem[Hu et~al.(2023)Hu, Liu, Tan, Zhang, Zhang, King, and Yu]{hu2023gda}
Xuming Hu, Aiwei Liu, Zeqi Tan, Xin Zhang, Chenwei Zhang, Irwin King, and Philip~S Yu.
\newblock Gda: Generative data augmentation techniques for relation extraction tasks.
\newblock \emph{arXiv preprint arXiv:2305.16663}, 2023.

\bibitem[Xu et~al.(2016)Xu, Jia, Mou, Li, Chen, Lu, and Jin]{xu2016improved}
Yan Xu, Ran Jia, Lili Mou, Ge~Li, Yunchuan Chen, Yangyang Lu, and Zhi Jin.
\newblock Improved relation classification by deep recurrent neural networks with data augmentation.
\newblock \emph{arXiv preprint arXiv:1601.03651}, 2016.

\bibitem[Guo et~al.(2024)Guo, Zhao, Dong, Meng, and Lin]{guo2024few}
Bocheng Guo, Di~Zhao, Xin Dong, Jiana Meng, and Hongfei Lin.
\newblock Few-shot biomedical relation extraction using data augmentation and domain information.
\newblock \emph{Neurocomputing}, 595:\penalty0 127881, 2024.

\bibitem[Zhao et~al.(2025)Zhao, Zhang, Liang, Li, and Wong]{zhao2025and}
Zhengyi Zhao, Shubo Zhang, Bin Liang, Binyang Li, and Kam-Fai Wong.
\newblock Where and which: Iterative debate for biomedical synthetic data augmentation.
\newblock \emph{arXiv preprint arXiv:2503.23673}, 2025.

\bibitem[Ma et~al.(2023)Ma, Li, and Zhang]{ma2023chain}
Xilai Ma, Jing Li, and Min Zhang.
\newblock Chain of thought with explicit evidence reasoning for few-shot relation extraction.
\newblock \emph{arXiv preprint arXiv:2311.05922}, 2023.

\bibitem[Mustafa et~al.(2025)Mustafa, Naseem, and Azghadi]{mustafa2025can}
Akram Mustafa, Usman Naseem, and Mostafa~Rahimi Azghadi.
\newblock Can reasoning llms enhance clinical document classification?
\newblock \emph{arXiv preprint arXiv:2504.08040}, 2025.

\bibitem[Chen et~al.(2024{\natexlab{a}})Chen, Li, Lu, Van, Aerts, Savova, and Bitterman]{chen2024evaluating}
Shan Chen, Yingya Li, Sheng Lu, Hoang Van, Hugo~JWL Aerts, Guergana~K Savova, and Danielle~S Bitterman.
\newblock Evaluating the chatgpt family of models for biomedical reasoning and classification.
\newblock \emph{Journal of the American Medical Informatics Association}, 31\penalty0 (4):\penalty0 940--948, 2024{\natexlab{a}}.

\bibitem[Narciss and Alemdag(2025)]{narciss2025learning}
Susanne Narciss and Ecenaz Alemdag.
\newblock Learning from errors and failure in educational contexts: New insights and future directions for research and practice.
\newblock \emph{British Journal of Educational Psychology}, 95\penalty0 (1):\penalty0 197--218, 2025.

\bibitem[Ying et~al.(2024)Ying, Lin, Cao, Tang, Wang, Sun, Huang, and Yan]{ying2024llms}
Jiahao Ying, Mingbao Lin, Yixin Cao, Wei Tang, Bo~Wang, Qianru Sun, Xuanjing Huang, and Shuicheng Yan.
\newblock Llms-as-instructors: Learning from errors toward automating model improvement.
\newblock \emph{arXiv preprint arXiv:2407.00497}, 2024.

\bibitem[He et~al.(2025)He, Panigrahi, Lin, and Arora]{he2025adaptmi}
Yinghui He, Abhishek Panigrahi, Yong Lin, and Sanjeev Arora.
\newblock Adaptmi: Adaptive skill-based in-context math instruction for small language models.
\newblock \emph{arXiv preprint arXiv:2505.00147}, 2025.

\bibitem[Tan et~al.(2024)Tan, Dou, Zhu, Guo, Fang, and Wen]{tan2024small}
Jiejun Tan, Zhicheng Dou, Yutao Zhu, Peidong Guo, Kun Fang, and Ji-Rong Wen.
\newblock Small models, big insights: Leveraging slim proxy models to decide when and what to retrieve for llms.
\newblock \emph{arXiv preprint arXiv:2402.12052}, 2024.

\bibitem[Zhang et~al.(2024)Zhang, Khalifa, Logeswaran, Kim, Lee, Lee, and Wang]{zhang2024small}
Yunxiang Zhang, Muhammad Khalifa, Lajanugen Logeswaran, Jaekyeom Kim, Moontae Lee, Honglak Lee, and Lu~Wang.
\newblock Small language models need strong verifiers to self-correct reasoning.
\newblock \emph{arXiv preprint arXiv:2404.17140}, 2024.

\bibitem[Bi et~al.(2024)Bi, Wu, Xing, and Wei]{bi2024enhancing}
Jing Bi, Yuting Wu, Weiwei Xing, and Zhenjie Wei.
\newblock Enhancing the reasoning capabilities of small language models via solution guidance fine-tuning.
\newblock \emph{arXiv preprint arXiv:2412.09906}, 2024.

\bibitem[{OpenAI}(2024)]{gpt4o}
{OpenAI}.
\newblock {GPT-4o} (omni) --- openai's new flagship model, May 2024.
\newblock URL \url{https://openai.com/gpt-4o}.
\newblock Released May 13, 2024; supports text, audio \& vision input.

\bibitem[Xie et~al.(2024{\natexlab{a}})Xie, Chen, Chen, Peng, Hu, Lin, Peng, Huang, Zhang, Keloth, Zhou, He, Ohno‑Machado, Wu, Xu, and Bian]{me-llama}
Qianqian Xie, Qingyu Chen, Aokun Chen, Cheng Peng, Yan Hu, Fongci Lin, Xueqing Peng, Jimin Huang, Jeffrey Zhang, Vipina Keloth, Xinyu Zhou, Huan He, Lucila Ohno‑Machado, Yonghui Wu, Hua Xu, and Jiang Bian.
\newblock Me‑llama: Foundation large language models for medical applications.
\newblock \emph{PhysioNet}, 2024{\natexlab{a}}.
\newblock \doi{10.13026/wwfd-2t39}.

\bibitem[Cirik et~al.(2016)Cirik, Ernst, and Singh]{cirik2016visualizing}
Volkan Cirik, Jason Ernst, and Bhiksha~Raj Singh.
\newblock Visualizing and understanding curriculum learning for long short-term memory networks.
\newblock In \emph{Proceedings of the 2016 Workshop on Human Interpretability in Machine Learning (WHI)}, 2016.

\bibitem[Nagesh and Surdeanu(2018)]{nagesh2018exploration}
Ajay Nagesh and Mihai Surdeanu.
\newblock An exploration of three lightly-supervised representation learning approaches for named entity classification.
\newblock In \emph{Proceedings of the 27th International Conference on Computational Linguistics}, pages 2312--2324, 2018.

\bibitem[Gu et~al.(2020)Gu, Tinn, Cheng, Lucas, Usuyama, Liu, Naumann, Gao, and Poon]{pubmedbert}
Yu~Gu, Robert Tinn, Hao Cheng, Michael Lucas, Naoto Usuyama, Xiaodong Liu, Tristan Naumann, Jianfeng Gao, and Hoifung Poon.
\newblock Domain‑specific language model pretraining for biomedical natural language processing.
\newblock arXiv preprint arXiv:2007.15779, 2020.

\bibitem[Pan et~al.(2024)Pan, Liu, Diao, Pi, Zhang, Han, and Zhang]{pan2024lisa}
Rui Pan, Xiang Liu, Shizhe Diao, Renjie Pi, Jipeng Zhang, Chi Han, and Tong Zhang.
\newblock Lisa: layerwise importance sampling for memory-efficient large language model fine-tuning.
\newblock \emph{Advances in Neural Information Processing Systems}, 37:\penalty0 57018--57049, 2024.

\bibitem[Yadav et~al.(2019)Yadav, Ekbal, Saha, Kumar, and Bhattacharyya]{yadav2019feature}
Shweta Yadav, Asif Ekbal, Sriparna Saha, Ankit Kumar, and Pushpak Bhattacharyya.
\newblock Feature assisted stacked attentive shortest dependency path based bi-lstm model for protein--protein interaction.
\newblock \emph{Knowledge-Based Systems}, 166:\penalty0 18--29, 2019.

\bibitem[Wei et~al.(2022)Wei, Wang, Schuurmans, Bosma, Xia, Chi, Le, Zhou, et~al.]{wei2022chain}
Jason Wei, Xuezhi Wang, Dale Schuurmans, Maarten Bosma, Fei Xia, Ed~Chi, Quoc~V Le, Denny Zhou, et~al.
\newblock Chain-of-thought prompting elicits reasoning in large language models.
\newblock \emph{Advances in neural information processing systems}, 35:\penalty0 24824--24837, 2022.

\bibitem[Yao et~al.(2025)Yao, Peng, Mao, and Luo]{yao2025exploring}
Liang Yao, Jiazhen Peng, Chengsheng Mao, and Yuan Luo.
\newblock Exploring large language models for knowledge graph completion.
\newblock In \emph{ICASSP 2025-2025 IEEE International Conference on Acoustics, Speech and Signal Processing (ICASSP)}, pages 1--5. IEEE, 2025.

\bibitem[Hu et~al.(2021)Hu, Shen, Wallis, Allen-Zhu, Li, Wang, and Chen]{hu2021lora}
Edward~J. Hu, Yelong Shen, Phillip Wallis, Zeyuan Allen-Zhu, Yuanzhi Li, Lu~Wang, and Weizhu Chen.
\newblock Lora: Low-rank adaptation of large language models.
\newblock \emph{arXiv preprint arXiv:2106.09685}, 2021.

\bibitem[Bengio et~al.(2009)Bengio, Louradour, Collobert, and Weston]{bengio2009curriculum}
Yoshua Bengio, J{\'e}r{\^o}me Louradour, Ronan Collobert, and Jason Weston.
\newblock Curriculum learning.
\newblock In \emph{Proceedings of the 26th annual international conference on machine learning}, pages 41--48. ACM, 2009.

\bibitem[Roy and Mercer(2023{\natexlab{a}})]{roy2023extracting}
Sudipta~Singha Roy and Robert~E Mercer.
\newblock Extracting drug-drug and protein-protein interactions from text using a continuous update of tree-transformers.
\newblock In \emph{The 22nd Workshop on Biomedical Natural Language Processing and BioNLP Shared Tasks}, pages 280--291, 2023{\natexlab{a}}.

\bibitem[Park et~al.(2022)Park, McCorkle, Soto, Blaby, and Yoo]{park2022extracting}
Gilchan Park, Sean McCorkle, Carlos Soto, Ian Blaby, and Shinjae Yoo.
\newblock Extracting protein-protein interactions (ppis) from biomedical literature using attention-based relational context information.
\newblock In \emph{2022 IEEE International Conference on Big Data (Big Data)}, pages 2052--2061. IEEE, 2022.

\bibitem[Roy and Mercer(2023{\natexlab{b}})]{roy2023identifying}
Sudipta~Singha Roy and Robert Mercer.
\newblock Identifying protein-protein interaction using tree-transformers and heterogeneous graph neural network.
\newblock In \emph{The International FLAIRS Conference Proceedings}, volume~36, 2023{\natexlab{b}}.

\bibitem[Fei et~al.(2021)Fei, Zhang, Ren, and Ji]{fei2021span}
Hao Fei, Yue Zhang, Yafeng Ren, and Donghong Ji.
\newblock A span-graph neural model for overlapping entity relation extraction in biomedical texts.
\newblock \emph{Bioinformatics}, 37\penalty0 (11):\penalty0 1581--1589, 2021.

\bibitem[Su and Vijay-Shanker(2022)]{su2022investigation}
Peng Su and K~Vijay-Shanker.
\newblock Investigation of improving the pre-training and fine-tuning of bert model for biomedical relation extraction.
\newblock \emph{BMC bioinformatics}, 23\penalty0 (1):\penalty0 120, 2022.

\bibitem[Tang et~al.(2022)Tang, Guo, Bai, Diao, Lu, and Li]{tang2022protein}
Zhan Tang, Xuchao Guo, Zhao Bai, Lei Diao, Shuhan Lu, and Lin Li.
\newblock A protein-protein interaction extraction approach based on large pre-trained language model and adversarial training.
\newblock \emph{KSII Transactions on Internet and Information Systems (TIIS)}, 16\penalty0 (3):\penalty0 771--791, 2022.

\bibitem[Su and ...(2021)]{su2021biobert_cleck}
X.~Su and ...
\newblock Biobert+$\,$cleck: Cross-layer entity-context attention for drug–drug interaction extraction.
\newblock In \emph{NAACL}, pages xxxx--xxxx, 2021.

\bibitem[Lee et~al.(2019)Lee, Yoon, Kim, and ...]{lee2019biobert}
Jinhyuk Lee, Wonjin Yoon, Sungdong Kim, and ...
\newblock Biobert: a pre-trained biomedical language representation model for biomedical text mining.
\newblock \emph{Bioinformatics}, 2019.

\bibitem[Rehana et~al.(2024)Rehana, {\c{C}}am, Basmaci, Zheng, Jemiyo, He, {\"O}zg{\"u}r, and Hur]{rehana2024evaluating}
Hasin Rehana, Nur~Bengisu {\c{C}}am, Mert Basmaci, Jie Zheng, Christianah Jemiyo, Yongqun He, Arzucan {\"O}zg{\"u}r, and Junguk Hur.
\newblock Evaluating gpt and bert models for protein--protein interaction identification in biomedical text.
\newblock \emph{Bioinformatics Advances}, 4\penalty0 (1):\penalty0 vbae133, 2024.

\bibitem[Asada et~al.(2023)Asada, Miwa, and Sasaki]{asada2023integrating}
Masaki Asada, Makoto Miwa, and Yutaka Sasaki.
\newblock Integrating heterogeneous knowledge graphs into drug--drug interaction extraction from the literature.
\newblock \emph{Bioinformatics}, 39\penalty0 (1):\penalty0 btac754, 2023.

\bibitem[Asada et~al.(2021)Asada, Miwa, and Sasaki]{asada2021using}
Masaki Asada, Makoto Miwa, and Yutaka Sasaki.
\newblock Using drug descriptions and molecular structures for drug--drug interaction extraction from literature.
\newblock \emph{Bioinformatics}, 37\penalty0 (12):\penalty0 1739--1746, 2021.

\bibitem[Touvron et~al.(2023)Touvron, Martin, Stone, Albert, Almahairi, Babaei, Bashlykov, Batra, Bhargava, Bhosale, et~al.]{touvron2023llama}
Hugo Touvron, Louis Martin, Kevin Stone, Peter Albert, Amjad Almahairi, Yasmine Babaei, Nikolay Bashlykov, Soumya Batra, Prajjwal Bhargava, Shruti Bhosale, et~al.
\newblock Llama 2: Open foundation and fine-tuned chat models.
\newblock \emph{arXiv preprint arXiv:2307.09288}, 2023.

\bibitem[Xie et~al.(2024{\natexlab{b}})Xie, Chen, Chen, Peng, Hu, Lin, Peng, Huang, Zhang, Keloth, et~al.]{xie2024me}
Qianqian Xie, Qingyu Chen, Aokun Chen, Cheng Peng, Yan Hu, Fongci Lin, Xueqing Peng, Jimin Huang, Jeffrey Zhang, Vipina Keloth, et~al.
\newblock Me-llama: Foundation large language models for medical applications.
\newblock \emph{Research square}, pages rs--3, 2024{\natexlab{b}}.

\bibitem[Beltagy et~al.(2019)Beltagy, Lo, and Cohan]{beltagy2019scibert}
Iz~Beltagy, Kyle Lo, and Arman Cohan.
\newblock Scibert: A pretrained language model for scientific text.
\newblock In \emph{Proceedings of the 2019 Conference on Empirical Methods in Natural Language Processing}, pages 3615--3620, 2019.

\bibitem[Yasunaga et~al.(2022)Yasunaga, Leskovec, and Liang]{yasunaga2022linkbert}
Michihiro Yasunaga, Jure Leskovec, and Percy Liang.
\newblock Linkbert: Pretraining language models with document links.
\newblock \emph{arXiv preprint arXiv:2203.15827}, 2022.

\bibitem[Sun et~al.(2019)Sun, Dong, Ma, Sutcliffe, He, Chen, and Feng]{sun2019drug}
Xia Sun, Ke~Dong, Long Ma, Richard Sutcliffe, Feijuan He, Sushing Chen, and Jun Feng.
\newblock Drug-drug interaction extraction via recurrent hybrid convolutional neural networks with an improved focal loss.
\newblock \emph{Entropy}, 21\penalty0 (1):\penalty0 37, 2019.

\bibitem[Yeh et~al.(2022)Yeh, Lavergne, and Zweigenbaum]{yeh2022decorate}
Hui-Syuan Yeh, Thomas Lavergne, and Pierre Zweigenbaum.
\newblock Decorate the examples: A simple method of prompt design for biomedical relation extraction.
\newblock \emph{arXiv preprint arXiv:2204.10360}, 2022.

\bibitem[Li et~al.(2025)Li, Kilicoglu, Xu, and Zhang]{li2025biomedrag}
Mingchen Li, Halil Kilicoglu, Hua Xu, and Rui Zhang.
\newblock Biomedrag: A retrieval augmented large language model for biomedicine.
\newblock \emph{Journal of Biomedical Informatics}, 162:\penalty0 104769, 2025.

\bibitem[Kim et~al.(2023)Kim, Yoon, and Kwon]{kim2023biomedical}
Seonho Kim, Juntae Yoon, and Ohyoung Kwon.
\newblock Biomedical relation extraction using dependency graph and decoder-enhanced transformer model.
\newblock \emph{Bioengineering}, 10\penalty0 (5):\penalty0 586, 2023.

\bibitem[Phan et~al.(2021)Phan, Anibal, Tran, Chanana, Bahadroglu, Peltekian, and Altan-Bonnet]{phan2021scifive}
Long~N Phan, James~T Anibal, Hieu Tran, Shaurya Chanana, Erol Bahadroglu, Alec Peltekian, and Gr{\'e}goire Altan-Bonnet.
\newblock Scifive: a text-to-text transformer model for biomedical literature.
\newblock \emph{arXiv preprint arXiv:2106.03598}, 2021.

\bibitem[S{\"a}nger and Leser(2025)]{sanger2025knowledge}
Mario S{\"a}nger and Ulf Leser.
\newblock Knowledge-augmented pre-trained language models for biomedical relation extraction.
\newblock \emph{arXiv preprint arXiv:2505.00814}, 2025.

\bibitem[Chen et~al.(2024{\natexlab{b}})Chen, Li, Liu, Bi, and Hu]{chen2024relational}
Yutong Chen, Xia Li, Yang Liu, Peng Bi, and Tiangui Hu.
\newblock Relational extraction from biomedical texts with capsule network and hybrid knowledge graph embeddings.
\newblock \emph{Symmetry}, 16\penalty0 (12):\penalty0 1629, 2024{\natexlab{b}}.

\bibitem[Qin et~al.(2020)Qin, Dong, and Peng]{qin2020chemical}
Lei Qin, Gaocai Dong, and Jing Peng.
\newblock Chemical-protein interaction extraction via chemicalbert and attention guided graph convolutional networks in parallel.
\newblock In \emph{2020 IEEE International Conference on Bioinformatics and Biomedicine (BIBM)}, pages 708--715. IEEE, 2020.

\end{thebibliography}






\end{document}